\documentclass[letterpaper, 10 pt, conference]{ieeeconf}  

\IEEEoverridecommandlockouts                              

\usepackage{hyperref}
\hypersetup{
    colorlinks=true,
    linkcolor=black,    
    urlcolor=blue,
    }
\usepackage{graphicx}
\usepackage{subcaption}
\usepackage{booktabs}
\usepackage{epsfig} 
\usepackage{amsmath} 
\usepackage{amssymb}  
\usepackage{mathrsfs}
\usepackage{cleveref}
\usepackage{color}
\usepackage{calc}
\usepackage{dsfont}
\usepackage{bm}
\usepackage{float}
\usepackage{url}
\usepackage{xspace}
\usepackage[dvipsnames]{xcolor}

\def \FigureAbbreviaition {Fig.}
\newcommand{\figref}[1]{\FigureAbbreviaition\ \ref{#1}}

\usepackage{amsfonts} 
\title{\LARGE \bf Trajectory Optimization for In-Hand Manipulation with Tactile Force Control}

\author{Haegu Lee, Yitaek Kim, Victor Melbye Staven, and Christoffer Sloth 
\thanks{Authors are with the Maersk Mc-Kinney Moller Institute, University of Southern Denmark, Denmark {\tt\small \{haeg, yik,vims,chsl\}@mmmi.sdu.dk}}
}

\usepackage{tikz}

\newcommand\submittedtext{%
  \footnotesize \textcopyright \text{ }2025 IEEE.  Personal use of this material is permitted.  Permission from IEEE must be obtained for all other uses, in any current or future media, including reprinting/republishing this material for advertising or promotional purposes, creating new collective works, for resale or redistribution to servers or lists, or reuse of any copyrighted component of this work in other works.}
\newcommand\submittednotice{%
\begin{tikzpicture}[remember picture,overlay]
\node[anchor=south,yshift=10pt] at (current page.south) {\fbox{\parbox{\dimexpr\textwidth-\fboxsep-\fboxrule\relax}{\submittedtext}}};
\end{tikzpicture}%
}

\begin{document}
\maketitle
\submittednotice

\thispagestyle{empty}
\pagestyle{empty}


\begin{abstract}
The strength of the human hand lies in its ability to manipulate objects precisely and robustly. In contrast, simple robotic grippers have low dexterity and fail to handle small objects effectively. This is why many automation tasks remain unsolved by robots.
This paper presents an optimization-based framework for in-hand manipulation with a robotic hand equipped with compact Magnetic Tactile Sensors (MTSs). 
We formulate a trajectory optimization problem using Nonlinear Programming (NLP) for finger movements while ensuring contact points to change along the geometry of the fingers. Using the optimized trajectory from the solver, we implement and test an open-loop controller for rolling motion. To further enhance robustness and accuracy, we introduce a force controller for the fingers and a state estimator for the object utilizing MTSs. The proposed framework is validated through comparative experiments, showing that incorporating the force control with compliance consideration improves the accuracy and robustness of the rolling motion. Rolling an object with the force controller is 30\% more likely to succeed than running an open-loop controller. The demonstration video is available at \url{https://youtu.be/6J_muL_AyE8}.
\end{abstract}

\section{Introduction}\label{sec:introduction}
In-hand manipulation skills are in many real-world applications such as industrial assembly, which inspires the significant development of robotic motion planning strategies. Recent research in robotics has actively investigated planning of intricate, human-like hand motions \cite{Jihong2020}, \cite{wilson2023cable}, \cite{Manes2024}. Yet, robotic in-hand motion planning with an object still remains challenging due to the complexity of contact dynamics, high dimensionality of the configuration space, and hardware limitations such as bulky vision-based tactile sensors \cite{she2021cable} and fingers \cite{Nikhil2014SimpleGripper}. 


Among various motion primitives, rolling an object between two fingers serves as a representative example of complex in-hand manipulation. This primitive has gained attention in robotics due to its crucial role in industry applications, such as screwing bolts \cite{Tang2024Screw} or inserting low-volume parts \cite{Azulay2023RAL}.  To achieve the rolling motion, the framework for learning contact dynamics with vision-based tactile feedback is proposed to roll a rigid long stick between two fingers, enhancing grasp robustness \cite{du2024stick}. From \cite{Zhang2022Rolling}, the adaptive rolling motion is achieved to the object's width by estimating the contact area from three binarized images using two visuotactile sensors attached to a parallel gripper. In the perspective of hardware solutions, various specialized gripper designs have been proposed \cite{Chen2014IROS}, \cite{Zuo2021}, \cite{Yan2022ICMA} to accomplish robust and efficient rolling motion. However, it is still non-trivial to achieve robust and adaptive rolling motion with high precision and success rate, even if the motion seems straightforward for humans.

\begin{figure}[t]
    \centering
    \includegraphics[width=1\linewidth]{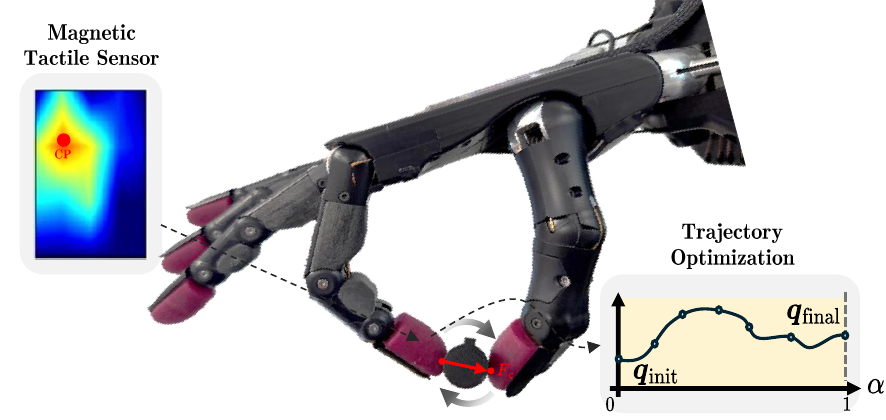}
    \caption{\small{Rolling an object using a robotic hand. The First finger, with 4 degrees of freedom (DoF), and the thumb, with 5 DoF, are equipped with Magnetic Tactile Sensors (MTSs).}}
    \vspace{-0.3cm}
    \label{fig:hero}
\end{figure}

In this vein, we utilize a commercial multi-finger robotic hand that is lightweight, compact, and highly human-like. Consequently, this robotic hand can provide the most similar in-hand manipulation skills to those of humans. Additionally, a magnetic tactile sensor shown in \figref{fig:hero}, is lighter and more compact than visuotactile sensors \cite{yuan2017gelsight}, while still being able to detect contact forces. Finally, an efficient and straightforward trajectory optimization method can be employed to generate rolling motion primitives while ensuring a firm grasp on the object.


 \begin{figure*}[t]
    \centerline{\includegraphics[width=1\linewidth]{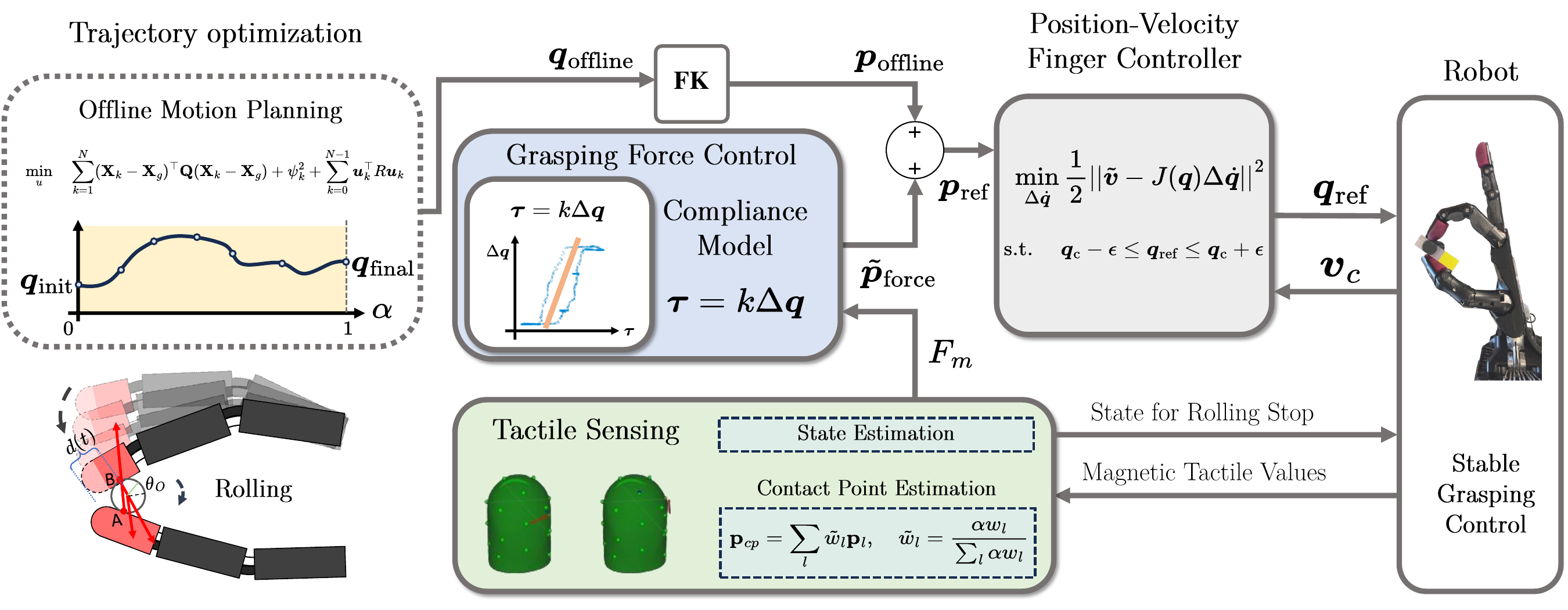}}
    \caption{\small{The architecture of the proposed framework. The model-based trajectory optimization generates offline finger reference motions for rolling, and the finger force controller tracks these motions using tactile-based state estimation for rolling. This ensures robust and stable grasping of a given object. Furthermore, the compliant behavior of the hand compromises the performance of the finger controller; thus, our framework also includes compliance compensation, which enhances the performance of the reference tracking control.}}
    \vspace{-0.3cm}
    \label{fig:framework} 
\end{figure*}

\subsection{Contributions}
As shown in \figref{fig:hero}, this paper aims to introduce a comprehensive framework that achieves sophisticated rolling motion for a human-like multi-finger robotic hand. The main contributions of this paper are described as follows:
\begin{itemize}
    \item Design a trajectory optimization for contact dynamics to generate multi-finger motion that continuously changes contact points along the fingers.
    \item Design a finger controller to track the motion reference from trajectory optimization while ensuring stable contact force using magnetic-based tactile perception. 
    \item Estimate the stiffness of the tendon-driven finger and compensate for compliant behaviors to improve the finger controller.
    \item Demonstrate the proposed framework through experimental evaluations on the real robotic setup.
\end{itemize}

This paper is organized as follows. Section~\ref{sec:related_works} presents related work to our research and section~\ref{sec:proposed_framework} introduces the components and overall structure of the proposed framework. In section ~\ref{sec:Results}, the proposed framework is verified with a real robotic hand. Finally, Section~\ref{sec:conclusions} concludes the paper and discusses future work.
\section{Related Work}\label{sec:related_works}
\subsection{In-Hand Manipulation}
To perform diverse and delicate tasks, in-hand manipulation is essential. In-hand manipulation can be achieved either by dexterous hands or by utilizing external forces, which leads to extrinsic dexterity. Without complex dynamic modeling, simple primitives that consider external forces can adjust the position or orientation of objects \cite{Nikhil2014SimpleGripper}. Modeling the relationships among the external environment, the gripper, and the object allows for more accurate attainment of the desired pose of the object \cite{mordatch2012contact, chavan2015prehensile, chavan2020planar, hou2018fast}. While frictional forces are taken into account, the method remains limited as it execute open-loop actions without predicting the actual pose of the object. Learning-based methods have also been studied for planning dexterous manipulation \cite{nagabandi2020deep, kumar2016optimal} but still do not estimate state of the objects that can be used for closing the control loop. \cite{andrychowicz2020learning} proposed a reinforcement learning method to learn dexterous in-hand manipulation with Shadow Dexterous hand. They successfully rotate a given object to the goal configuration. However their approach relied on multiple vision sensors, and the object was relatively large.

 The limitations in pose estimation of grasped objects can be addressed with recently developed tactile sensors \cite{yuan2017gelsight, lambeta2020digit, taylor2022gelslim}. Due to the flexibility of hardware reconfiguration in parallel grippers, in-hand manipulation using tactile sensors has been extensively studied \cite{hogan2020tactile, kim2023simultaneous, bronars2024texterity}. However, since two-finger gripper have a limited degree of the freedom, the external environment is often utilized to generate dexterous hand manipulation. Incorporating tactile sensors into a human-like multi-finger hand is challenging, as tactile sensors are often larger than fingertips. Therefore, various control methods have been explored to have an effective finger control for robotic hands to enhance the stable and precise in-hand manipulation.
 
\subsection{Finger Control}
Finger control can be achieved in several ways, tailored to the specific control objectives. The impedance control scheme in torque-controlled robotic fingers is employed in \cite{ButterfassDLRHand2001} and extended with learning-based control parameter estimation to enhance the performance of the controller \cite{Sievers2022}. The finger dynamics model can be used to implement inverse dynamics control for finger control \cite{Xue2023IROS}, based on a joint torque controller. Besides these control methods, hybrid position/force control \cite{YOSHIKAWA2010199}, force-position control \cite{ROJASGARCIA2022233},  force-velocity control \cite{Li-RSS-13} and position-velocity control \cite{Raković2018} provide alternative control designs for fingers to deal with intricate interactions between fingers and the contact environments.

\section{Proposed Framework}\label{sec:proposed_framework}
In this section, we introduce the proposed framework for accurate and robust in-hand manipulation with tactile sensing. The framework consists of offline motion planning for the rolling motion, a  state estimation for the object pose, a finger motion and force controller with compliance compensation of the fingers as shown in \figref{fig:framework}. In the following, we present detailed explanations of each component in our framework.

 \begin{figure}[h]
    \centering
    \includegraphics[width=1\linewidth]{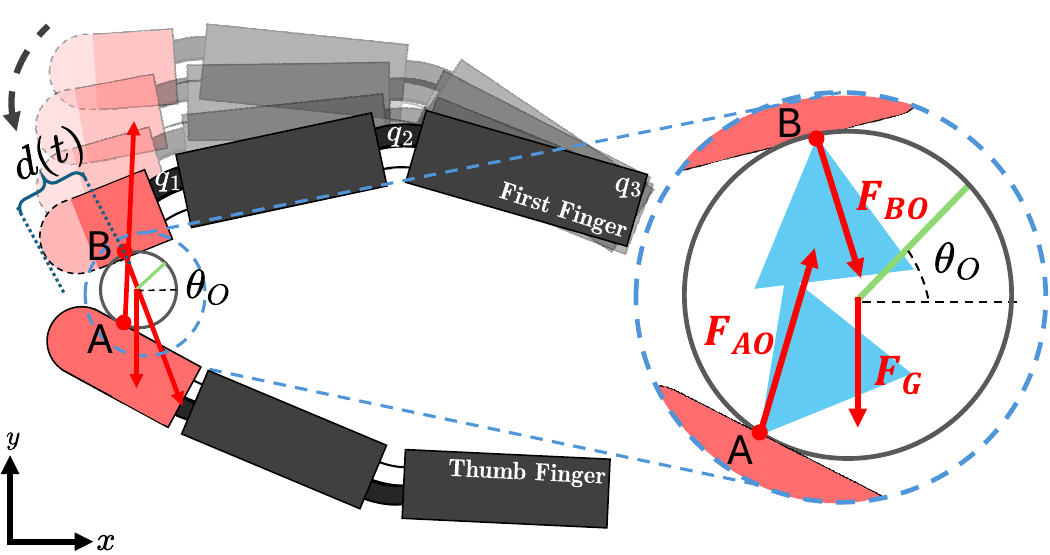}
    \caption{\small{Illustration of the rolling motion of a cylinder manipulated by two fingers.  The upper finger represents the First finger and the bottom finger represents the Thumb finger.}}
    \vspace{-0.3cm}
    \label{fig:mechanics} 
\end{figure}
\subsection{Contact Dynamics for Rolling Motion}
Consider two fingers in contact with an object as shown in \figref{fig:mechanics}. The forces exerted onto the object at the contact points $A$ and $B$ are called $\mathbf{F}_{AO}$ and $\mathbf{F}_{BO}$. The following describes the quasi-static dynamics of the system similar to \cite{shirai2023tactile}.
To enable rolling, sustained contact must be obtained between both fingers and the object. Therefore, we ensure that the contact dynamics holds the quasi-static equilibrium of the object between the fingers as follows:
\begin{gather}
    \mathbf{F}_{AO}+\mathbf{F}_{BO}+\mathbf{F}_{G} = 0  \label{eq:force_equilibrium}\\
    \bm{\tau}_{AO}+\bm{\tau}_{BO} +\bm{\tau}_{G}=0  \label{eq:torque_equilibrium}
\end{gather}
where \eqref{eq:force_equilibrium} and \eqref{eq:torque_equilibrium} represent the static equilibrium force and moment on the object. 
In practical implementations, the contact between a finger and an object is more accurately represented as patch contact rather than point contact. While the fingertip of the Shadow Hand undergoes slight deformation upon contact, accurately modeling patch contact and incorporating deformation remains a significant challenge in real-world applications. Therefore, the contact model is simplified as a point contact for offline motion planning. To prevent the object from sliding instead of rolling, contact forces are constrained within the friction cone according to Coulomb's law. The contact forces consist of normal and tangential compoents, expressed as \(\mathbf{F}_{AO}=[f_{n}^A, \bm{f}_{t}^A]^\top\), \(\mathbf{F}_{BO}=[f_{n}^B, \bm{f}_{t}^B]^\top\), which satisfy Coulomb’s law as follows:
\begin{gather}
- \mu_i \|\bm{f}_t^i\| \leq f_n^i \leq \mu_i \|\bm{f}_t^i\|, \quad f_n^i \geq 0, \quad \forall i \in \{A, B\}
\end{gather}
where $\|\bm{f}_{t}^i\|$ denotes the 2-norm of $\bm{f}_{t}^i$ at contact point $i \in \{A, B\}$ and \(\mu_i\) is the coefficient of friction between the fingertip and the object. By approximating the contact as a point contact, moment friction is neglected. 

Rolling an object requires a continuous transition of contact points while maintaining persistent contact. Contact points A and B must transition along the fingers to achieve proper rolling. However, contact-rich manipulation tasks are usually performed with stationary contact. To generate a trajectory that allows contact point B to change, we add a virtual prismatic joint to Finger B. As shown in \figref{fig:mechanics}, $d(t)$ corresponds to the length of the virtual prismatic link, which changes as the object rolls between the fingers to maintain contact. Contact between the virtual prismatic link and the object is enforced by: 
 \begin{gather}
d_{k-1} - d_{k} - r\cdot(\theta_{o,k-1}-\theta_{o,k}) = 0
\label{eq:prismatic_constraint}
\end{gather}
where $\theta_O$ represents the orientation of the object and $k$ denotes a time step for the trajectory and $r$ is the radius of object. Under the no-slip assumption, the state of the object can be represented in a manner similar to \eqref{eq:prismatic_constraint}.
 \begin{gather}
\| \mathbf{c}_{k-1} - \mathbf{c}_k \| - r \cdot(\theta_{o,k-1} - \theta_{o}) = 0
\label{eq:obj_kinematics_constraint}
\end{gather}
where \( \mathbf{c}_k = [x_{o, k}, y_{o, k}]^\top \) represents the center position of the rolling object at time step \( k \).
\subsection{Offline Motion Planning}
Trajectory optimization is used to plan an offline rolling motion reference for the first finger. To simplify the problem, the thumb is considered flat as the possible travel distance for the object is relatively small. The contact model is formulated as a complementarity problem alongside trajectory optimization, incorporating complementarity constraints as presented in \cite{posa2014direct}. We assume that the object remains in quasi-static equilibrium, considering the simplified planar dynamics. The optimization is performed in discrete time, with a horizon of $N$ time steps. 
The trajectory optimization for rolling an object is formulated as follows:
\begin{subequations}
\begin{align}
    \min_{\mathbf{u}_k} \quad & \sum_{k=1}^{N} (\mathbf{X}_k - \mathbf{X}_g)^\top \mathbf{Q}(\mathbf{X}_k - \mathbf{X}_g) + \psi_k^2 +\sum_{k=0}^{N-1} \bm{u}_k^\top R \bm{u}_k \label{eq:cost} \\
    \text{s.t.} \quad & (1), (2), (3), (4), (5) \label{eq:constraints}
\end{align}
\label{eq:TO}%
\end{subequations}
where $\mathbf{X}_k = [x_{\textnormal{o},k}, y_{\textnormal{o},k},\theta_{\textnormal{o},k}]^\top$ represents the state of the object at time step $k$, consisting of its position $(x_{\textnormal{o},k}, y_{\textnormal{o},k})$ and orientation $\theta_{o,k}$. The goal state is denoted by $\mathbf{X}_g = [x_{\textnormal{g}}, y_{\textnormal{g}},\theta_{\textnormal{g}}]^\top$, which defines the goal position and orientation for the object. The control input for the fingers is denoted as $\mathbf{u}_k$ and is a vector of joint torques. The weight matrices $\mathbf{Q}$ and $\mathbf{R}$ are positive semi-definite and positive definite, respectively. 

The term $\psi_k$ represents the distance between the fingers and the object, is added to the cost function to enforce contact between the fingers and the object. Incorporation of $\psi$ in the cost function is crucial as it ensures that fingers reach the object and apply force while maintaining contact. As shown in \figref{fig:mechanics}, the initial configuration of the first finger is not in contact with the object.

Another important constraint is \eqref{eq:prismatic_constraint}, which defines a virtual prismatic joint that enables rolling while allowing the contact point between the finger and the object to change continuously. If constraint \eqref{eq:prismatic_constraint} is not included in the optimization problem, the solver is likely to fail to find a solution. In such case, the finger may make contact only at the fingertip, resulting in the object being pushed by the fingertip rather than naturally following the geometry of the fingers during rolling.
The nonlinear problem \eqref{eq:TO} was solved using IPOPT \cite{wachter2006implementation} in CasADi \cite{andersson2019casadi}. The solution consists of fingertip positions and force information, which serve as the reference trajectory for the rolling motion.

\subsection{State Estimation with Tactile Sensing}
Unlike vision-based tactile sensors, Magnetic Tactile Sensors (MTSs) respond to changes in pressure rather than using visual information. As a result, estimation problems become difficult to approach using image processing methods, such as Principal Component Analysis. To address this, a state estimation method using 17 MTSs is proposed to estimate the pose of an object for accurate rolling along a given reference trajectory. 

\begin{figure}[h]
    \centering
    \includegraphics[width=0.9\linewidth]{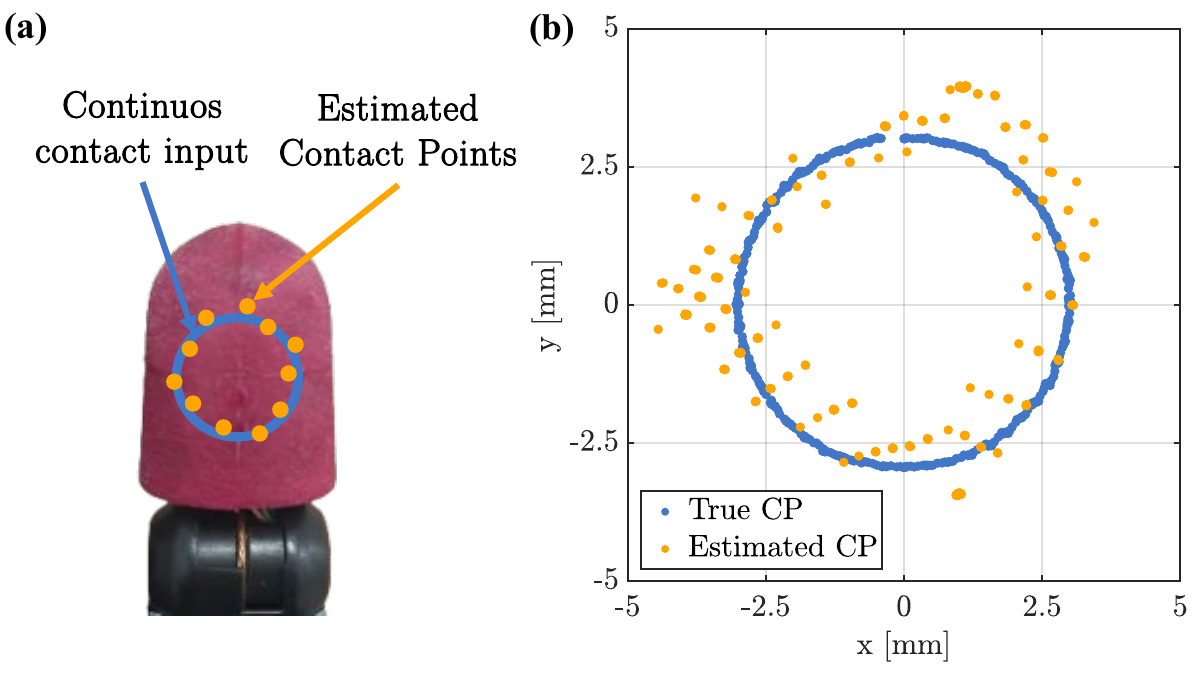}
    \caption{\small{Illustration of the contact point estimator with continuous contact input.}}
    \vspace{-0.3cm}
    \label{fig:cop_estimator} 
\end{figure}

\subsubsection{Contact Point Estimation}
Let  $\bm{f}_i, \bm{p}_i \in\mathbb{R}^3$  be the magnetic sensor values and the sensor positions at each finger, respectively, where $i \in I=\{ 1,\dots,n_s\}$ and $n_s$ represents the total number of sensors attached to the fingers. Define $\textbf{s}\in \mathbb{R}^{n_s}$ as a vector with elements $s_i=||\bm{f}_i||$, where $s_i$ is the force magnitude at point $\bm{p}_i$. The vector $\textbf{s}$ is normalized to a range from $0$ to $255$ as follows:
\begin{equation}
s^{'}_{i} = 255 \bigg( \frac{s_{i} - \min(\textbf{s})}{\max(\textbf{s}) - \min(\textbf{s})}\bigg),
\label{eq.normalize}
\end{equation}
where $s^{'}_{i}$ is a normalized element, and $\min(\cdot), \max(\cdot)$ return the minimum and maximum magnitudes of $s$, respectively. From the normalized vector $s^{'}$, we select the element with the maximum magnitude  and denote it by $s_{i^*}$. Subsequently, we use a threshold, $T > 0$ to compute a weight.
\begin{equation}
W_{i} = \max \left( \frac{s^{'}_{i} - T}{s_{i^*}^{'} - T}, 0 \right),
\label{eq.max weight}
\end{equation}
Let $J \subseteq I$ be a set of indices corresponding to the $n_n$ nearest neighbors elements. Then,
the contact point is computed as:
\begin{equation}
    \bm{p}_{\textnormal{cp}} = \sum_{i\in J} \tilde{w}_{i} \bm{p}_i\textnormal{ ,} \quad \tilde{w}_{i} = \frac{s^{'}_{i}}{\sum_{j \in J} {s^{'}_{j}}}
\end{equation}
The estimated contact point is used to calculate the distance between the initial contact point and the measured contact point, and the geometry of the object is utilized to predict the rotation of the object, $\theta_o$. The estimation of the travel distance can be improved by incorporating the geometric information of the fingers. \figref{fig:cop_estimator} shows the estimated contact point while maintaining sustained contact between the finger surface and a UR10e equipped with a tool. The estimated contact points tend to exhibit noise along the true contact points, varying depending on the contact geometry. 
\begin{figure*}[t]
    \centering
    \includegraphics[width=1\linewidth]{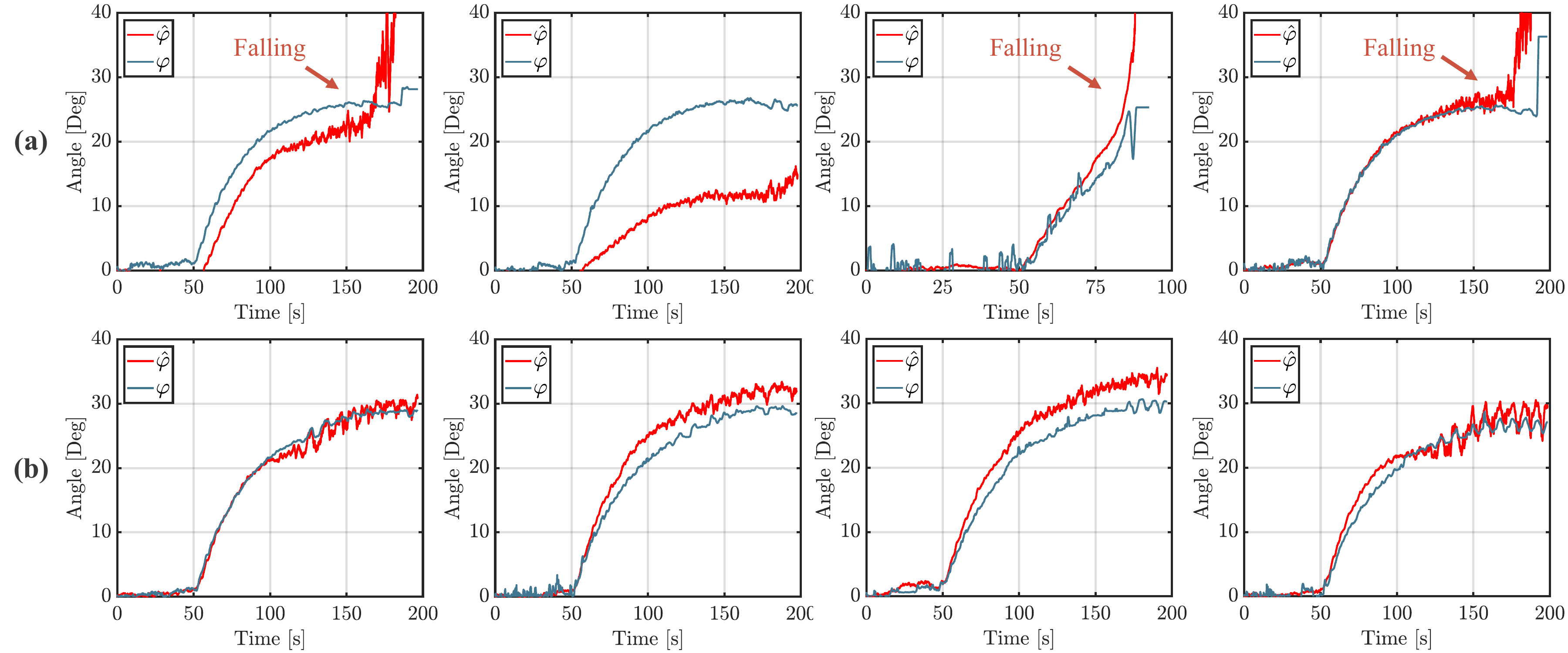}
    \caption{\small{Ground truth ($\varphi$) and estimated orientation ($\hat\varphi$) of the object during rolling motion. The row (a) of the figure shows the object's orientation under an open-loop controller, while the row (b) shows the orientation with an additional force controller.}}
    \vspace{-0.3cm}
    \label{Exp:graphs}
\end{figure*}

\subsection{Finger Motion and Force Control}\label{sec:finger_controller}
We design the finger controller based on a cascade position/velocity controller. Since the robotic hand includes coupled actuation between $q_1$ and $q_2$ in \figref{fig:mechanics}, it is non-trivial to control a fingertip through analytical inverse kinematic-based approach. Instead, a Jacobian-based velocity controller is employed to track the reference position trajectory of each fingertip. 

However, tracking a reference trajectory using only position and velocity control is insufficient for stable object manipulation. Since the manipulation task is simplified for trajectory optimization, an open-loop controller may fail to maintain a stable grasp during the rolling motion. While the intrinsic compliance of the hand aids stability, it cannot fully compensate for significant force variation. To address this issue, we incorporate a force controller to ensure robust manipulation. Additionally, a compliance analysis of tendons was conducted to account for finger deformations during the force control process.
\subsubsection{Finger Controller}
Let us define the position tracking error, $\tilde{\bm{p}} \in \mathbb{R}^3 $ as follows:
\begin{equation}
    \tilde{\bm{p}} = \bm{p}_{\textnormal{ref}} - \bm{p}_{\textnormal{c}},
\end{equation}
where $ \bm{p}_{\textnormal{ref}}\in \mathbb{R}^3$ and $ \bm{p}_{\textnormal{c}}\in \mathbb{R}^3$ are the reference and current measured positions of a fingertip, respectively. The position of the tip can be obtained from forward kinematics of the hand. Subsequently, we use a proportional–integral–derivative (PID) controller to define the reference velocity $\bm{v}_{\textnormal{ref}} \in \mathbb{R}^3$ 

In the same way, a PID controller is utilized to track $\bm{v}_{\textnormal{ref}}$ with the current $\bm{v}_{c} \in \mathbb{R}^3$ defined as:
\begin{equation}
     \bm{v}_{\textnormal{c}}= J(\bm{q})\dot{\bm{q}}, \label{velocity_rel}
\end{equation}
where $\bm{q} \in \mathbb{R}^m$ and $\dot{\bm{q}} \in \mathbb{R}^m$ are the current joint angle and angular velocity at each finger, respectively. $m$ indicates the number of degrees of freedom of each finger, and $J(\bm{q}) \in \mathbb{R}^{m\times3}$ is the Jacobian matrix of each finger. Consequently, the final reference configurations at each finger are calculated as follows:
\begin{equation}
    \bm{q}_{\textnormal{ref}} = \bm{q}_{\textnormal{c}} + \Delta\dot{\bm{q}}\textnormal{dt}, \label{pro:final_ctrl_law}
\end{equation}
\begin{equation}
    \Delta\dot{\bm{q}} =J^+(\bm{q}) \Big(\underbrace{k_{\textrm{pv}}\tilde{\bm{v}}_{\textnormal{e}} + k_{\textrm{iv}}\int\tilde{\bm{v}}_{\textnormal{e}}\textnormal{dt}   + k_{\textrm{dv}}\dot{\tilde{\bm{v}}}_{\textnormal{e}}}_{\tilde{\bm{v}}} \Big), 
    \label{pro:jac_ctrl}
\end{equation}
where $\tilde{\bm{v}}_{\textnormal{e}} = \bm{v}_{\textnormal{ref}} - \bm{v}_{\textnormal{c}}$, and $k_{\textrm{pv}},k_{\textrm{iv}},k_{\textrm{dv}} \in \mathbb{R}$ are PID gains for the velocity controller. In \eqref{pro:jac_ctrl}, $\Delta\dot{\bm{q}}$ is determined based on the pseudo-inverse matrix, which may cause $\bm{q}_{\textnormal{ref}}$ to deviate significantly from $\bm{q}_{\textnormal{c}}$, leading to large changes in the controller output. To resolve this, we enforce a constraint ensuring that $\bm{q}_{\textnormal{ref}}$ remain close to the current configuration of each finger. Consequently, we formulate the following linear least-squares problem:
\begin{align}
    &\min_{\Delta\dot{\bm{q}}} \frac{1}{2} {\vert\vert \tilde{\bm{v}} - J(\bm{q})\Delta\dot{\bm{q}}\vert\vert}^2 \label{eq:qp_pvcontroller}\\
    \text{s.t. } \quad &\bm{q}_{\textnormal{c} }- \epsilon \leq \bm{q}_{\textnormal{ref}} \leq \bm{q}_{\textnormal{c}} + \epsilon \nonumber
\end{align}
where $\epsilon \in \mathbb{R}^r$ represents an admissible bound for each joint.
\subsubsection{Force Controller}
Kinematic model and simulations of the Shadow Hand do exist \cite{menagerie2022github}; however, this model does not include the coupling between individual finger joints and their compliance when external forces are applied. Consequently, we provide such a model and show how to integrate the coupling mode into the force controller.

\begin{figure}[h]
    \centering
    \includegraphics[width=1\linewidth]{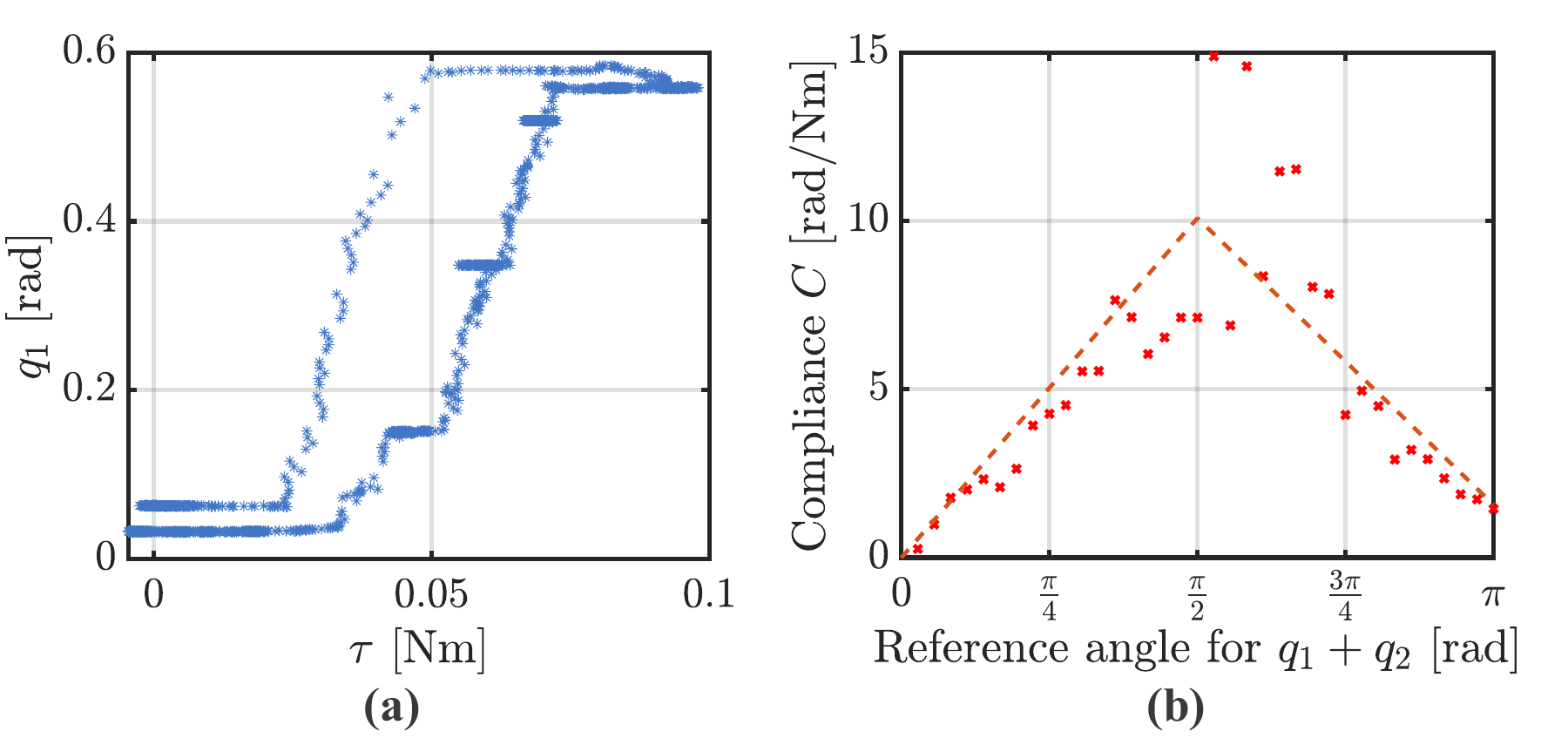}
    \caption{\small{ (a) Change in $q_1$ during a single trial with varying external torque $\tau$. (b) Compliance of the first finger under various finger configurations.}}
    \vspace{-0.3cm}
    \label{fig:compliance} 
\end{figure}

\figref{fig:compliance}(a) shows $q_1$ of the first finger when a varying force is applied at the fingertip, while \figref{fig:compliance}(b) indicates the variation in compliance depending on the reference angle for $q_1 + q_2$. It is seen that the compliance increases linearly from $0$ to $\frac{\pi}{2}$ and then decreases linearly after $\frac{\pi}{2}$. This trend implies that the mechanical structure of the finger exhibits variable compliance, which should be considered when performing force control with the finger.

The 17 MTSs provide magnetic sensor values that respond to the force applied to the finger. These values can be interpreted as the contact force magnitude, ${F}_m \in \mathbb{R}$, which is obtained by averaging the magnitudes of all individual sensors. The desired and current force vectors $\bm{F}_{\textnormal{des}}\in \mathbb{R}^3$ and $\bm{F}_{\textnormal{c}}\in \mathbb{R}^3$ are defined as follows: 
\begin{align}
\bm{F}_{\textnormal{des}} = F_{\textnormal{des}} \frac{\bm{p_n}}{\|\bm{p_n}\|},&&
\bm{F}_{\textnormal{c}}\ = F_m\frac{\bm{p_n}}{\|\bm{p_n}\|} &&
\bm{p_n} = \bm{p}_\textnormal{cp,ff} - \bm{p}_\textnormal{cp,th}
\end{align}



where  $\bm{p}_{\textnormal{cp,ff}} \in \mathbb{R}^3$,  $\bm{p}_{\textnormal{cp,th}}\in \mathbb{R}^3$ represent the estimated contact points of the first finger and thumb, respectively, and $F_{\textnormal{des}} \in \mathbb{R}$ denotes the desired force magnitude. With given force vectors, the desired position command is generated as follows:
\begin{align}
    \tilde{\bm{p}}_\textnormal{force}= {k_{\textrm{pf}}\tilde{\bm{F}}_{\textnormal{e}}} + k_{\textrm{if}}\int\tilde{\bm{F}}_{\textnormal{e}}\textnormal{dt} + C \tilde{\bm{F}}_{\textnormal{e}}
\end{align}
where \(  \tilde{\bm{p}}_\textnormal{force} \) is the desired position from the force controller, \( C \) represents the compliance, \( \tilde{\bm{F}}_{\textnormal{e}} = \bm{F}_{\textnormal{des}} - \bm{F}_c \) and $k_{\textrm{pf}},k_{\textrm{if}} \in \mathbb{R}$ are PI gains for the force controller, respectively.

\section{Experimental Results}\label{sec:Results}
In this section, we present the experiments conducted to evaluate the proposed framework and discuss the results. First, we show the outcome of executing the trajectory generated through trajectory optimization with an open-loop controller. We then introduce a force controller integrated with the open-loop controller, analyze the resulting changes in the desired position, and examine its impact on the rolling motion.
\subsection{Experiment Setup}
To validate the rolled angle of a grasped object, an ArUco marker was attached to a 3D-printed cylindrical object with a radius of 7.5 mm and used to obtain ground-truth pose data. The estimated rotation from the contact point was then compared with the ground truth. We conducted two sets of experiments. 

First, we executed only the open-loop reference trajectory generated offline. Then, we integrated a force controller with the offline reference trajectory to evaluate its impact on grasp stability and rolling accuracy.

\subsection{Open-Loop Controller}
 The optimal solution from \eqref{eq:TO} serves as the reference trajectory for rolling motion with two fingers. During the initial phase of trajectory following, the fingers successfully roll the object due to the inherent compliance, which contributes to a stable grasp. However, the fingers lose the object as the contact force between the fingers and the object gradually decreases when the object rolls down the thumb. This leads to a failed grasp and inaccuracies in state estimation of the object. \figref{Exp:graphs}(a) shows cases where the object is dropped due to a gradually decreasing force.
 
 We observed that the open-loop controller itself occasionally succeeds in rolling the object, although without accuracy, due to the natural compliance of the fingers. However the object was dropped in 6 out of 10 trials when executing only the reference trajectory. Even in the remaining 4 trials, as shown in the second graph in \figref{Exp:graphs}(a), accurately estimating the state of the object was challenging due to the insufficient applied force, which also resulted in a lack of robustness to disturbances.

 \begin{figure}[ht]
    \centering
    \includegraphics[width=1\linewidth]{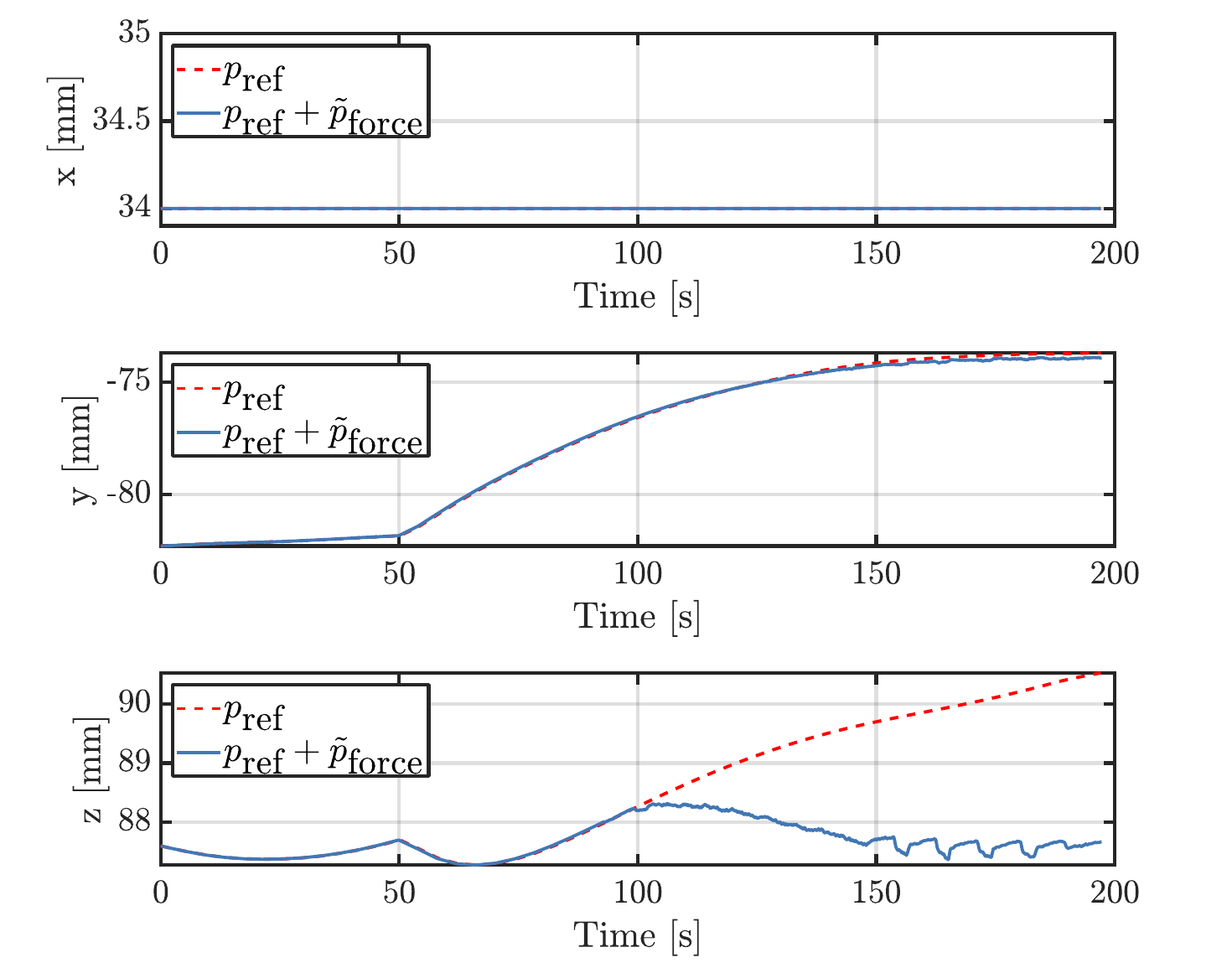}
    \caption{\small{Comparison of the reference position from offline motion planning and the refined reference position adjusted by the force controller.}}
    \vspace{-0.3cm}
    \label{fig:refined_ref_trjaecory}
\end{figure}

\subsection{Open-Loop and Force Controller}
\begin{figure}[ht]
    \vspace{-0.4cm}
    \centering
    \includegraphics[width=1\linewidth]{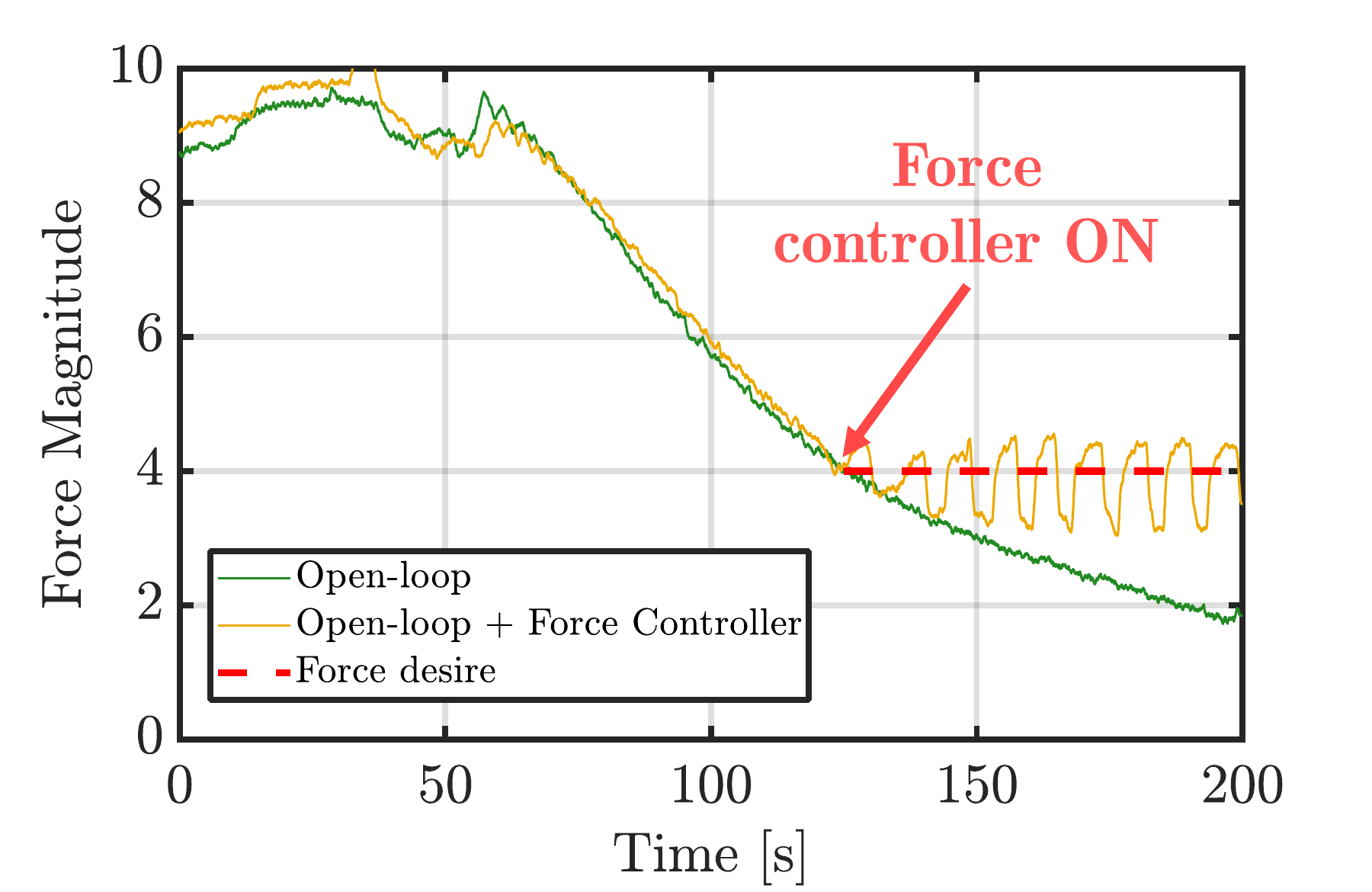}
    \caption{\small{Force magnitude during the rolling motion}}
    \vspace{-0.3cm}
    \label{fig:force_mag}
\end{figure}
As shown in \figref{fig:refined_ref_trjaecory}, the position-force control method adjusts the reference trajectory to ensure that the contact force remains at the desired level between the object and the fingers.
The force controller is activated when the current force magnitude from the first finger falls below a predefined threshold, indicating that the applied force is insufficient to stably grasp the object. The switching condition for the force controller exist only because the initial grasp provides sufficient magnitude as shown in \figref{fig:force_mag}. Once activated, the force controller generates position commands to ensure that the applied force remains within a predefined range, maintaining a stable grasp as the object rolls along the finger geometry. 

\figref{fig:force_mag} shows the changes in force magnitude from the MTSs on the first finger. Unlike the open-loop controller, where the force magnitude continuously decreases throughout the trajectory, integrating the force controller with the reference trajectory enables the force magnitude to remain close to the desired force magnitude. Although the force magnitude tends to oscillate. Since the force controller maintains a certain level of contact force, the state estimator becomes more stable compared to the open-loop case, increasing the likelihood of satisfying the terminal state conditions for the object. Rolling was successfully achieved in 7 out of 10 trials, while the remaining 3 trials exhibited deviations from the ground truth.

\section{Conclusion}\label{sec:conclusions}
This paper introduces a framework for in-hand manipulation with human-like robotic hands that are equipped with tactile sensors. We present the finger force controller with offline motion planning using a state estimator with tactile sensor and NLP. By demonstrating the rolling of an object with a radius of 7.5 mm using two fingers, we showed that the proposed method can perform robust and accurate rolling motion. As we use relatively compact tactile sensor compared to vision based tactile sensors, this research can be applied to tasks that require precise and robust manipulation of small objects in tight spaces. For future research, we plan to investigate the rolling of objects with complex or deformable shapes, where the dynamics are uncertain, and aim to achieve a higher success rate.

\section*{Acknowledgement}
This research was supported by Pioneer Center for Accelerating P2X Materials Discovery (CAPeX), DNRF grant number P3 and Fabrikant Vilhelm Pedersen og Hustrus Legat.

\bibliographystyle{IEEEtran}
\bibliography{reference}

@ARTICLE{Jihong2020,
  author={Zhu, Jihong and Navarro, Benjamin and Passama, Robin and Fraisse, Philippe and Crosnier, André and Cherubini, Andrea},
  journal={IEEE Robotics and Automation Letters}, 
  title={Robotic Manipulation Planning for Shaping Deformable Linear Objects WithEnvironmental Contacts}, 
  year={2020},
  volume={5},
  number={1},
  pages={16-23},
  doi={10.1109/LRA.2019.2944304}}

@INPROCEEDINGS{wilson2023cable,
  author={Wilson, Achu and Jiang, Helen and Lian, Wenzhao and Yuan, Wenzhen},
  booktitle={IEEE Int. Conf. Robot. Autom. }, 
  title={Cable Routing and Assembly using Tactile-driven Motion Primitives}, 
  year={2023},
  volume={},
  number={},
  pages={10408-10414},
  doi={10.1109/ICRA48891.2023.10161069}}

@article{she2021cable,
  title={Cable manipulation with a tactile-reactive gripper},
  author={She, Yu and Wang, Shaoxiong and Dong, Siyuan and Sunil, Neha and Rodriguez, Alberto and Adelson, Edward},
  journal={Int. J. Robot. Res.},
  volume={40},
  number={12-14},
  pages={1385--1401},
  year={2021},
  publisher={SAGE Publications Sage UK: London, England}
}

@Article{Manes2024,
author={Manes, Lupo
and Fichera, Sebastiano
and Fakhruldeen, Hatem
and Cooper, Andrew I.
and Paoletti, Paolo},
title={A soft cable loop based gripper for robotic automation of chemistry},
journal={Scientific Reports},
year={2024},
month={Apr},
day={17},
volume={14},
number={1},
pages={8899},
issn={2045-2322},
doi={10.1038/s41598-024-59372-1},
}

@article{yuan2017gelsight,
  title={Gel{S}ight: High-resolution robot tactile sensors for estimating geometry and force},
  author={Yuan, Wenzhen and Dong, Siyuan and Adelson, Edward H},
  journal={Sensors},
  volume={17},
  number={12},
  pages={2762},
  year={2017},
  publisher={MDPI}
}

@article{lambeta2020digit,
  title={{DIGIT}: A novel design for a low-cost compact high-resolution tactile sensor with application to in-hand manipulation},
  author={Lambeta, Mike and Chou, Po-Wei and Tian, Stephen and Yang, Brian and Maloon, Benjamin and Most, Victoria Rose and Stroud, Dave and Santos, Raymond and Byagowi, Ahmad and Kammerer, Gregg and others},
  journal={IEEE Robotics and Automation Letters},
  volume={5},
  number={3},
  pages={3838--3845},
  year={2020},
  publisher={IEEE}
}

@inproceedings{taylor2022gelslim,
  title={Gel{S}lim 3.0: High-resolution measurement of shape, force and slip in a compact tactile-sensing finger},
  author={Taylor, Ian H and Dong, Siyuan and Rodriguez, Alberto},
  booktitle={2022 International Conference on Robotics and Automation (ICRA)},
  pages={10781--10787},
  year={2022},
  organization={IEEE}
}

@INPROCEEDINGS{Zhang2022Rolling,
  author={Zhang, Wuyi and Xia, Chongkun and Zhu, Xiaojun and Liu, Houde and Liang, Bin},
  booktitle={2022 IEEE International Conference on Systems, Man, and Cybernetics (SMC)}, 
  title={Tac{R}ot: A Parallel-Jaw Gripper with Rotatable Tactile Sensors for In-Hand Manipulation}, 
  year={2022},
  volume={},
  number={},
  pages={423-429},
  doi={10.1109/SMC53654.2022.9945388}}

@inproceedings{du2024stick,
  title={Stick Roller: Precise In-hand Stick Rolling with a Sample-Efficient Tactile Model},
  author={Du, Yipai and Zhou, Pokuang and Wang, Michael Yu and Lian, Wenzhao and She, Yu},
  booktitle={2024 IEEE/RSJ International Conference on Intelligent Robots and Systems (IROS)},
  pages={2312--2318},
  year={2024},
  organization={IEEE}
}

@INPROCEEDINGS{Tang2024Screw,
  author={Tang, Ling and Jia, Yan-Bin and Xue, Yuechuan},
  booktitle={2024 IEEE International Conference on Robotics and Automation (ICRA)}, 
  title={Robotic Manipulation of Hand Tools: The Case of Screwdriving}, 
  year={2024},
  volume={},
  number={},
  pages={13883-13890},
  doi={10.1109/ICRA57147.2024.10610831}}

@INPROCEEDINGS{Nikhil2014SimpleGripper,
  author={Dafle, Nikhil Chavan and Rodriguez, Alberto and Paolini, Robert and Tang, Bowei and Srinivasa, Siddhartha S. and Erdmann, Michael and Mason, Matthew T. and Lundberg, Ivan and Staab, Harald and Fuhlbrigge, Thomas},
  booktitle={2014 IEEE International Conference on Robotics and Automation (ICRA)}, 
  title={Extrinsic dexterity: In-hand manipulation with external forces}, 
  year={2014},
  volume={},
  number={},
  pages={1578-1585},
  doi={10.1109/ICRA.2014.6907062}}

@inproceedings{hogan2020tactile,
  title={Tactile dexterity: Manipulation primitives with tactile feedback},
  author={Hogan, Francois R and Ballester, Jose and Dong, Siyuan and Rodriguez, Alberto},
  booktitle={2020 IEEE international conference on robotics and automation (ICRA)},
  pages={8863--8869},
  year={2020},
  organization={IEEE}
}

@inproceedings{kim2023simultaneous,
  title={Simultaneous tactile estimation and control of extrinsic contact},
  author={Kim, Sangwoon and Jha, Devesh K and Romeres, Diego and Patre, Parag and Rodriguez, Alberto},
  booktitle={2023 IEEE International Conference on Robotics and Automation (ICRA)},
  pages={12563--12569},
  year={2023},
  organization={IEEE}
}

@inproceedings{bronars2024texterity,
  title={{TEX}terity: Tactile Extrinsic deXterity},
  author={Bronars, Antonia and Kim, Sangwoon and Patre, Parag and Rodriguez, Alberto},
  booktitle={2024 IEEE International Conference on Robotics and Automation (ICRA)},
  pages={7976--7983},
  year={2024},
  organization={IEEE}
}

@ARTICLE{Azulay2023RAL,
  author={Azulay, Osher and Monastirsky, Maxim and Sintov, Avishai},
  journal={IEEE Robotics and Automation Letters}, 
  title={Haptic-Based and $SE(3)$-Aware Object Insertion Using Compliant Hands}, 
  year={2023},
  volume={8},
  number={1},
  pages={208-215},
  doi={10.1109/LRA.2022.3224670}}

@INPROCEEDINGS{Sievers2022,
  author={Sievers, Leon and Pitz, Johannes and Bäuml, Berthold},
  booktitle={2022 International Conference on Robotics and Automation (ICRA)}, 
  title={Learning Purely Tactile In-Hand Manipulation with a Torque-Controlled Hand}, 
  year={2022},
  volume={},
  number={},
  pages={2745-2751},
  doi={10.1109/ICRA46639.2022.9812093}}

@inproceedings{mordatch2012contact,
  title={Contact-invariant optimization for hand manipulation},
  author={Mordatch, Igor and Popovi{\'c}, Zoran and Todorov, Emanuel},
  booktitle={Proceedings of the ACM SIGGRAPH/Eurographics symposium on computer animation},
  pages={137--144},
  year={2012}
}

@inproceedings{chavan2015prehensile,
  title={Prehensile pushing: In-hand manipulation with push-primitives},
  author={Chavan-Dafle, Nikhil and Rodriguez, Alberto},
  booktitle={2015 IEEE/RSJ International Conference on Intelligent Robots and Systems (IROS)},
  pages={6215--6222},
  year={2015},
  organization={IEEE}
}

@article{chavan2020planar,
  title={Planar in-hand manipulation via motion cones},
  author={Chavan-Dafle, Nikhil and Holladay, Rachel and Rodriguez, Alberto},
  journal={The International Journal of Robotics Research},
  volume={39},
  number={2-3},
  pages={163--182},
  year={2020},
  publisher={SAGE Publications Sage UK: London, England}
}

@article{andrychowicz2020learning,
  title={Learning dexterous in-hand manipulation},
  author={Andrychowicz, OpenAI: Marcin and Baker, Bowen and Chociej, Maciek and Jozefowicz, Rafal and McGrew, Bob and Pachocki, Jakub and Petron, Arthur and Plappert, Matthias and Powell, Glenn and Ray, Alex and others},
  journal={The International Journal of Robotics Research},
  volume={39},
  number={1},
  pages={3--20},
  year={2020},
  publisher={SAGE Publications Sage UK: London, England}
}

@inproceedings{hou2018fast,
  title={Fast planning for 3{D} any-pose-reorienting using pivoting},
  author={Hou, Yifan and Jia, Zhenzhong and Mason, Matthew T},
  booktitle={2018 IEEE International Conference on Robotics and Automation (ICRA)},
  pages={1631--1638},
  year={2018},
  organization={IEEE}
}

@inproceedings{nagabandi2020deep,
  title={Deep dynamics models for learning dexterous manipulation},
  author={Nagabandi, Anusha and Konolige, Kurt and Levine, Sergey and Kumar, Vikash},
  booktitle={Conference on Robot Learning},
  pages={1101--1112},
  year={2020},
  organization={PMLR}
}

@inproceedings{kumar2016optimal,
  title={Optimal control with learned local models: Application to dexterous manipulation},
  author={Kumar, Vikash and Todorov, Emanuel and Levine, Sergey},
  booktitle={2016 IEEE International Conference on Robotics and Automation (ICRA)},
  pages={378--383},
  year={2016},
  organization={IEEE}
}

@misc{menagerie2022github,
  author = {Zakka, Kevin and Tassa, Yuval and {MuJoCo Menagerie Contributors}},
  title = {{MuJoCo Menagerie: A collection of high-quality simulation models for MuJoCo}},
  url = {http://github.com/googldeepmind/mujoco\_menagerie},
  year = {2022},
}

@INPROCEEDINGS{Xue2023IROS,
  author={Xue, Yuechuan and Tang, Ling and Jia, Yan-Bin},
  booktitle={2023 IEEE/RSJ International Conference on Intelligent Robots and Systems (IROS)}, 
  title={Dynamic Finger Gaits via Pivoting and Adapting Contact Forces}, 
  year={2023},
  volume={},
  number={},
  pages={8784-8791},
  doi={10.1109/IROS55552.2023.10342156}}

@article{YOSHIKAWA2010199,
title = {Multifingered robot hands: Control for grasping and manipulation},
journal = {Annual Reviews in Control},
volume = {34},
number = {2},
pages = {199-208},
year = {2010},
issn = {1367-5788},
doi = {https://doi.org/10.1016/j.arcontrol.2010.09.001},
author = {Tsuneo Yoshikawa}
}

@INPROCEEDINGS{Chen2014IROS,
  author={Chen, Fei and Cannella, Ferdinando and Canali, Carlo and D'Imperio, Mariapaola and Hauptman, Traveler and Sofia, Giuseppe and Caldwell, Darwin},
  booktitle={2014 IEEE/RSJ International Conference on Intelligent Robots and Systems}, 
  title={A study on data-driven in-hand twisting process using a novel dexterous robotic gripper for assembly automation}, 
  year={2014},
  volume={},
  number={},
  pages={4470-4475},
  doi={10.1109/IROS.2014.6943195}}

@Article{Zuo2021,
author={Zuo, Shiping
and Li, Jianfeng
and Dong, Mingjie},
title={Design, modeling, and manipulability evaluation of a novel four-DOF parallel gripper for dexterous in-hand manipulation},
journal={Journal of Mechanical Science and Technology},
year={2021},
month={Jul},
day={01},
volume={35},
number={7},
pages={3145-3160},
issn={1976-3824},
doi={10.1007/s12206-021-0636-7}
}

@INPROCEEDINGS{Yan2022ICMA,
  author={Yan, Yonggan and Guo, Shuxiang and Yang, Cheng and Lyu, Chuqiao and Zhang, Liuqing},
  booktitle={2022 IEEE International Conference on Mechatronics and Automation (ICMA)}, 
  title={The {PG2} {G}ripper: an Underactuated Two-fingered Gripper for Planar Manipulation}, 
  year={2022},
  volume={},
  number={},
  pages={680-685},
  doi={10.1109/ICMA54519.2022.9856375}}

@INPROCEEDINGS{ButterfassDLRHand2001,
  author={Butterfass, J. and Grebenstein, M. and Liu, H. and Hirzinger, G.},
  booktitle={Proceedings 2001 ICRA. IEEE International Conference on Robotics and Automation (Cat. No.01CH37164)}, 
  title={{DLR}-{Hand} {II}: next generation of a dextrous robot hand}, 
  year={2001},
  volume={1},
  number={},
  pages={109-114 vol.1},
  doi={10.1109/ROBOT.2001.932538}}

@article{ROJASGARCIA2022233,
title = {Force/position control with bounded actions on a dexterous robotic hand with two-degree-of-freedom fingers},
journal = {Biocybernetics and Biomedical Engineering},
volume = {42},
number = {1},
pages = {233-246},
year = {2022},
issn = {0208-5216},
doi = {https://doi.org/10.1016/j.bbe.2021.12.006},
author = {Lina N. Rojas-García and César A. Chávez-Olivares and Isela Bonilla-Gutiérrez and Marco O. Mendoza-Gutiérrez and Fernando Ramírez-Cardona}
}

@Article{Raković2018,
author={Rakovi{\'{c}}, Mirko
and Anil, Govind
and Mihajlovi{\'{c}}, {\v{Z}}ivorad
and Savi{\'{c}}, Sr{\dj}jan
and Naik, Siddhata
and Borovac, Branislav
and Gottscheber, Achim},
title={Fuzzy position-velocity control of underactuated finger of {FTN} robot hand},
journal={Journal of Intelligent {\&} Fuzzy Systems},
year={2018},
publisher={IOS Press},
volume={34},
pages={2723-2736},
note={4},
issn={1875-8967},
doi={10.3233/JIFS-17879}
}

@INPROCEEDINGS{Li-RSS-13, 
    AUTHOR    = {Qiang Li AND Carsten Sch{\"u}rmann AND Robert Haschke AND Helge Ritter}, 
    TITLE     = {A Control Framework for Tactile Servoing}, 
    BOOKTITLE = {Proceedings of Robotics: Science and Systems}, 
    YEAR      = {2013}, 
    ADDRESS   = {Berlin, Germany}, 
    MONTH     = {June},
    DOI       = {10.15607/RSS.2013.IX.045} 
}

@inproceedings{shirai2023tactile,
  title={Tactile tool manipulation},
  author={Shirai, Yuki and Jha, Devesh K and Raghunathan, Arvind U and Hong, Dennis},
  booktitle={2023 IEEE International Conference on Robotics and Automation (ICRA)},
  pages={12597--12603},
  year={2023},
  organization={IEEE}
}

@article{wachter2006implementation,
  title={On the implementation of an interior-point filter line-search algorithm for large-scale nonlinear programming},
  author={W{\"a}chter, Andreas and Biegler, Lorenz T},
  journal={Mathematical programming},
  volume={106},
  pages={25--57},
  year={2006},
  publisher={Springer}
}

@article{andersson2019casadi,
  title={Cas{AD}i: a software framework for nonlinear optimization and optimal control},
  author={Andersson, Joel AE and Gillis, Joris and Horn, Greg and Rawlings, James B and Diehl, Moritz},
  journal={Mathematical Programming Computation},
  volume={11},
  pages={1--36},
  year={2019},
  publisher={Springer}
}

@article{posa2014direct,
  title={A direct method for trajectory optimization of rigid bodies through contact},
  author={Posa, Michael and Cantu, Cecilia and Tedrake, Russ},
  journal={The International Journal of Robotics Research},
  volume={33},
  number={1},
  pages={69--81},
  year={2014},
  publisher={Sage Publications Sage UK: London, England}
}
\end{document}